\documentclass[journal]{IEEEtran}
\usepackage{xcolor,soul,framed}
\colorlet{shadecolor}{yellow}
\usepackage{graphicx} 
\graphicspath{{../pdf/}{../jpeg/}}
\DeclareGraphicsExtensions{.pdf,.jpeg,.png}
\usepackage{array}
\usepackage{mdwtab}
\usepackage{eqparbox}
\usepackage{lipsum} 
\usepackage{makecell}
\usepackage{multirow}
\usepackage{multicol}
\usepackage{booktabs}
\usepackage{colortbl}
\usepackage{bm}
\usepackage{xcolor}
\usepackage{caption}
\usepackage{subcaption}
\usepackage{booktabs}
\usepackage{listings}
\usepackage{tabularx}
\usepackage{tikz}
\usepackage{rotating}
\usepackage{fontawesome5}
\usepackage{nicematrix}
\usepackage{ulem}
\usepackage{makecell}
\usepackage{amsmath}
\usepackage{amssymb}
\usepackage{hyperref}
\hypersetup{
    colorlinks=false,
    pdfborder={0 0 0},
    linkcolor=black,
    citecolor=black,
    urlcolor=black
}
\usepackage{url}
\begin{document}
\bstctlcite{IEEEexample:BSTcontrol}
    \title{U3M: Unbiased Multiscale Modal Fusion Model for Multimodal Semantic Segmentation}
  \author{Bingyu Li\(^\textbf{\textdagger}\),
      Da Zhang\(^\textbf{\textdagger}\),
      Zhiyuan Zhao,
      Junyu Gao,~\IEEEmembership{Member,~IEEE,} and
      Xuelong Li\(^*\),~\IEEEmembership{Fellow,~IEEE}
  \thanks{\textbf{\textdagger} : Equal Contribution; *Corresponding author: Xuelong Li.}

  \thanks{Bingyu Li, Zhiyuan Zhao and Xuelong Li are with the Institute of Artificial Intelligence (TeleAI), China Telecom, P. R. China. (E-mail: libingyu0205@163.com; tuzixini@163.com; xuelong\_li@ieee.org).}
  \thanks{Da Zhang and Junyu Gao are with the School of Artificial Intelligence, OPtics and ElectroNics (iOPEN), Northwestern Polytechnical University, Xi'an 710072, China and with the Institute of Artificial Intelligence (TeleAI), China Telecom, P. R. China. (E-mail: dazhang@mail.nwpu.edu.cn; gjy3035@gmail.com).}% <-this % stops a space
  % \thanks{Zhiyuan Zhao and Xuelong Li is with the Institute of Artificial Intelligence (TeleAI), China Telecom, P. R. China. (E-mail: tuzixini@163.com; xuelong\_li@ieee.org).}
  }

\markboth{IEEE TRANSACTIONS ON MULTIMEDIA}{Roberg \MakeLowercase{\textit{et al.}}: U3M: Unbiased Multiscale Modal Fusion Model for Multimodal Semantic Segmentation}

% ====================================================================
\maketitle

% === ABSTRACT ====================================================================
% =================================================================================
\begin{abstract}
%\boldmath
Multimodal semantic segmentation is a pivotal component of computer vision and typically surpasses unimodal methods by utilizing rich information set from various sources.
Current models frequently adopt modality-specific frameworks that inherently biases toward certain modalities.
Although these biases might be advantageous in specific situations, they generally limit the adaptability of the models across different multimodal contexts, thereby potentially impairing performance.
To address this issue, we leverage the inherent capabilities of the model itself to discover the optimal equilibrium in multimodal fusion and introduce U3M: An \underline{U}nbiased \underline{M}ultiscale \underline{M}odal Fusion \underline{M}odel for Multimodal Semantic Segmentation.   
Specifically, this method involves an unbiased integration of multimodal visual data. Additionally, we employ feature fusion at multiple scales to ensure the effective extraction and integration of both global and local features.   
Experimental results demonstrate that our approach achieves superior performance across multiple datasets, verifing its efficacy in enhancing the robustness and versatility of semantic segmentation in diverse settings.
Our code is available at \href{https://github.com/LiBingyu01/U3M-multimodal-semantic-segmentation}{\textcolor{blue}{U3M-multimodal-semantic-segmentation}}.

\end{abstract}

% === KEYWORDS ====================================================================
% =================================================================================

% \fi
\IEEEpeerreviewmaketitle

% ====================================================================
% ====================================================================
% ====================================================================

% === I. INTRODUCTION =============================================================
% =================================================================================
\section{Introduction}

\begin{figure}
\centering
\includegraphics[width=\linewidth]{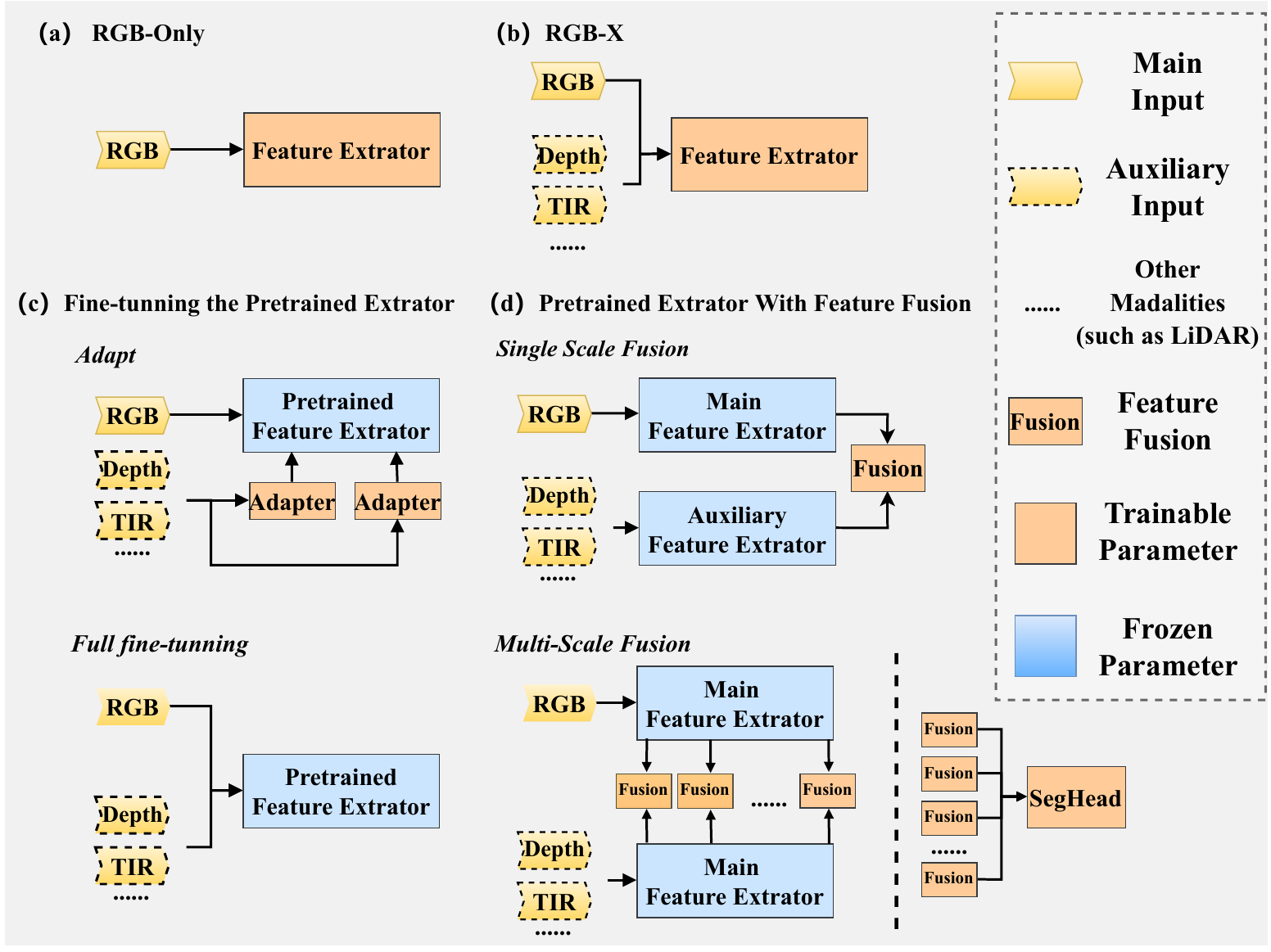}
\caption{The Evolution of Multimodal Semantic Segmentation Model Architectures. (a) Training a  feature extractor using only RGB images. (b) Sharing a trainable feature extractor between RGB images and other modalities. (c) Up: Sharing a fine-tunable pre-trained feature extractor between RGB images and other modalities. Down: Fine-tuning adapters for different modalities with one frozen feature extrator. (d) Up: Single-scale feature fusion within the model. Down: Multiscale feature fusion.}
\label{fig:TU1}
\vspace{-10pt}
\end{figure}

\IEEEPARstart{S}{emantic} segmentation \cite{chen2017deeplab, chen2018encoder,xie2021segformer} is a crucial task within the field of computer vision, with applications spanning various domains such as scene understanding \cite{liao2022kitti, dong2023egfnet, hazirbas2017fusenet, zhou2023cacfnet}, autonomous driving \cite{fu2019dual, liu2023multi}, etc. 
RGB-based semantic segmentation (depicted Fig. \ref{fig:TU1}(a)), serving as a foundational task, is suitable for the analysis of most scenes and has been extensively explored by the research community, yielding many impressive works and experimental results \cite{long2015fully, jin2021mining, zhao2017pyramid}. 
However, employing only RGB channels to segment certain complex and special scenes presents challenges, particularly where RGB information is elusive \cite{zhang2023cmx, dong2023egfnet}. 
In contrast to RGB cameras, which depend on visible light and often falter in darkness, thermal infrared (TIR) sensors directly detect heat emissions from objects, offering substantial contrast in the absence of light \cite{liu2023multi, zhou2023cacfnet}. Additionally, other modalities such as depth \cite{hazirbas2017fusenet, hu2019acnet} and LiDAR \cite{zhang2023delivering, liang2022multimodal} can also provide additional visual semantic information and are increasingly being integrated into semantic segmentation efforts, as illustrated in Fig. \ref{fig:TU1}(b).

In the context of multimodal semantic segmentation, multimodal feature extraction, serving as the foundation for downstream tasks, has been extensively explored in numerous studies \cite{chen2017deeplab, long2015fully}. Although these models have demonstrated exceptional performance in certain scenarios and datasets, they are typically trained from scratch on specific small-scale datasets, the limited number of images and scenes in multimodal visual training datasets leads to poor generalization across different scenes and data, not to mention better feature extraction capabilities. In contrast to aforementioned models trained from scratch, recent pretrained models, mostly based on Convolutional Neural Network (CNN) \cite{he2016deep} and Transformer\cite{vaswani2017attention}, have shown strong performance on downstream tasks by building models with large-scale parameters and pretraining on vast amounts of visual images. The rise of large-scale visual pretraining models has led to a series of studies exploring those applications in semantic segmentation. For instance, a large-scale pretrained convolutional neural network\cite{he2016deep}, demonstrated excellent semantic segmentation capabilities and performance by pretraining on ImageNet. Building on this well-established pretrained model, \cite{hu2019acnet} introduced ACNet, which refines feature extraction through asymmetric convolution blocks, enhancing network accuracy and robustness. Compared with CNN-based pretrained models, Transformers offer a larger receptive field \cite{vaswani2017attention} and stronger global modeling capabilities. Recent research \cite{Liu_2021_ICCV, dosovitskiy2020image} using plain vision Transformers for information extraction struggles with multiscale semantic information. To address this, Transformers with moving and multiscale windows have been explored \cite{dong2022cswin, yuan2021hrformer}. Multiscale information extractors have proven more practical. To balance speed and accuracy, we use the hierarchical multiscale vision Transformer Segformer \cite{xie2021segformer} as the modal information extractor.

Although extracted by large-scale pretrained models with strong visual information extraction capabilities, visual multimodal semantic information still presents some domain gaps. To bridge this, novel fine-tuning strategies using prompting techniques have been proposed \cite{he2024prompting}, efficiently adapting pretrained models to multimodal modes with minimal updates. Another approach involves using pretrained models as modality extractors, followed by fine-tuning with limited multimodal data, as shown in Fig. \ref{fig:TU1}(c). However, this often leads to catastrophic forgetting. Therefore, designing an additional feature fusion module is receiving increasing attention. For example, \cite{fu2019dual} designed an additional dual attention network to complement the pretrained feature extractor at the only low-resolution feature map (Fig. \ref{fig:TU1}(d)), which does not affect the original pretrained model. However, using single-scale information presents problems such as weak local spatial information and low input resolution. To address these issues, FuseNet \cite{hazirbas2017fusenet} integrates multiscale RGB-Depth multimodal features at every encoder stage, effectively improving performance on the SUN RGB-D benchmark. Similarly, CACFNet \cite{zhou2023cacfnet} incorporates a cross-modal attention fusion module that extracts multiscale RGB-thermal information. Besides, lots of extensive research \cite{zhang2023delivering, kaykobad2023multimodal} have demonstrated multiscale modal fusion modules (as shown in Fig. \ref{fig:TU1}(d)) enhance the adaptive integration of multimodal. 

\begin{figure}
\centering
\includegraphics[width=\linewidth]{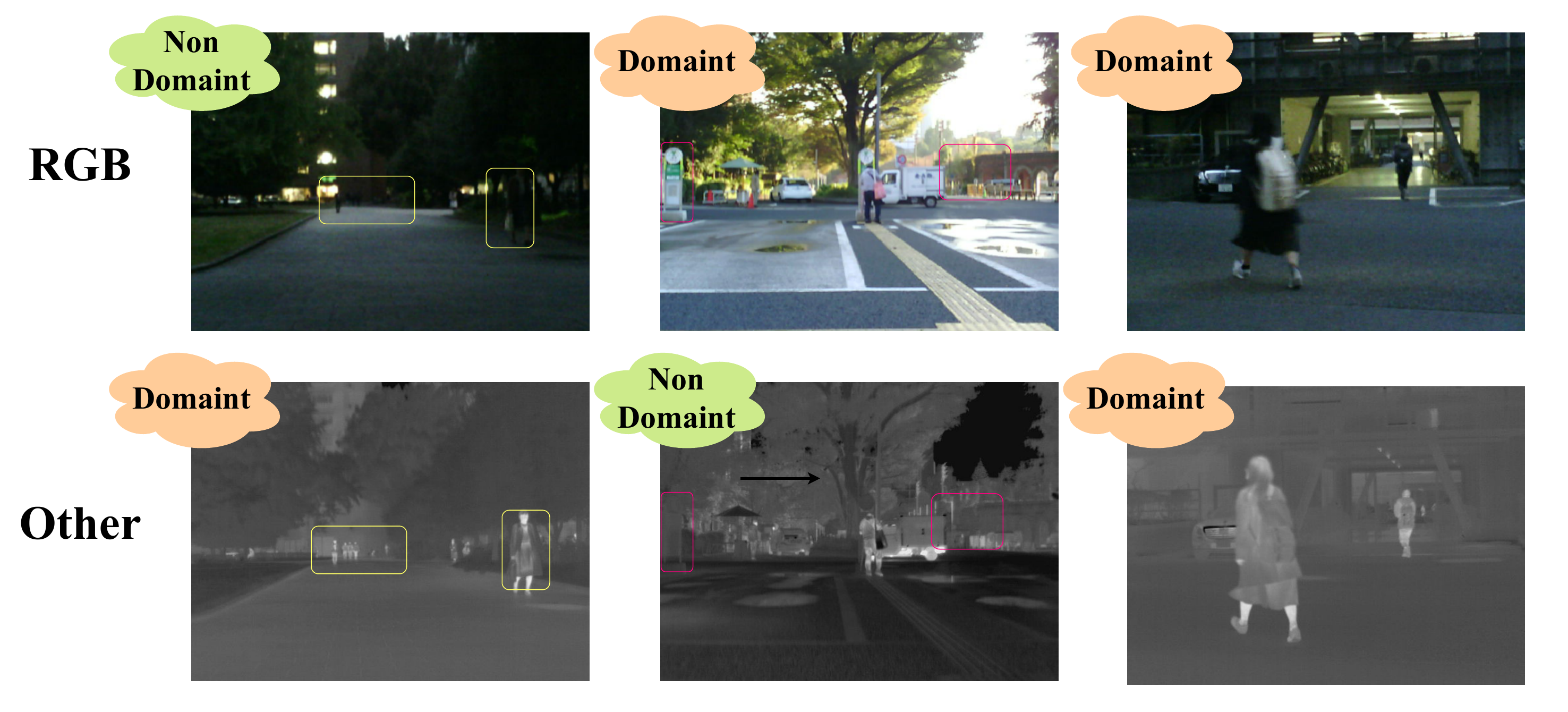}
\caption{The dynamic dominant correlation of multimodal data in different scenes. Left: Under conditions of insufficient light, infrared images can capture more intricate details than RGB images. Middle: In outdoor situations where light is abundant, infrared images tend to lose more details, while RGB images showcase their superiority. Right: In certain common instances, the detailed information in infrared images and RGB images can serve as a complement to each other.}
\label{fig:TU2}
\vspace{-10pt}
\end{figure}

Despite the excellent results achieved by existing work, particularly \cite{zhang2023cmx, kaykobad2023multimodal}, most of the multimodal fusions discussed above, notably CMNEXT \cite{zhang2023delivering}, exhibit some modalities bias. Specifically, these methods typically prioritize one modality as dominant and treat others as auxiliary (generally in the format of RGB+Xs) \cite{cao2024bi}. However, these approaches overlook the dynamic dominant correlation within multimodal data \cite{cao2024bi, reza2023multimodal}, which hampers the ability to fully utilize complementary multimodal information in complex scenarios (as illustrated in Fig.\ref{fig:TU2}), thus limiting performance. 

To avoid the inherent modal bias in model design and facilitate the fusion of multi-scale modal information, we first introduce an Unbiased Multiscale Modal Fusion Model (U3M) by treating all modalities equitably and performing multiscale fusion, which is different from the existing technology. This strategy enables the model to autonomously generate modal preferences applicable to various segmentation scenarios. Secondly, based on the fusion paradigm, we develop two multiscale fusion modules utilizing multiscale pooling and convolution, which effectively integrate and fuse global and local information across different scales for multimodal information. Finally, the model’s efficacy is validated by outstanding results across multiple datasets. Our contributions are summarized as follows:

\begin{itemize}
   \item We demonstrate the presence of modal bias in model design and develop an unbiased modal fusion methodology. This approach leverages the inherent properties of the model to autonomously generate modality preferences, substantially reducing biases introduced by manual design interventions.
    \item We design a multiscale model fusion layer that incorporates multiscale convolution and pooling. In this way, the fusion layer enhances the capability for multiscale modal fusion throughout various feature encoding stages.
    \item 
    Comprehensive experiments are conducted on two challenging multimodel semantic segmentation benchmarks (i.e., FMB \cite{liu2023multi}, Mcubes \cite{liang2022multimodal}), which further demonstrates that our method achieves state-of-the-art performance.
    
\end{itemize}

% === \section{Related Works}
% ========================
% =================================================================================
\section{Related Works}
\subsection{Semantic Segmentation}
The field of semantic segmentation has seen significant advancements, particularly with the advent of fully convolutional networks that revolutionized pixel classification \cite{long2015fully}. Based on this, some improvements such as multi-scale feature extraction and fusion have proven effective \cite{chen2017deeplab, hou2020strip}. For instance, Chen et al. further developed DeepLab by integrating an encoder-decoder architecture for efficient multi-scale context aggregation \cite{chen2018encoder}. Similarly, Zhao et al. introduced a pyramid pooling module to aggregate context from varied regions at multiple scales \cite{zhao2017pyramid}.

Channel and self-attention mechanisms have evolved to capture more global semantic information. Lin et al. developed RefineNet, employing multi-path refinement networks with channel attention for high-resolution segmentation \cite{lin2017refinenet}. Choi et al. proposed CARS, emphasizing channel-wise attention for region-based segmentation \cite{choi2020cars}. Another paradigm, Context-based refinement algorithms, integrates extensive background contextual information \cite{jin2021mining, zhang2018context}. Some other context-based models leverage context to refine segmentation through adaptive feature recalibration \cite{yu2020context}, while \cite{yuan2021ocnet} developed OCNet to enhance semantic understanding by aggregating object context.

Edge detection techniques also serve as complementary cues for semantic segmentation. For example, Li et al. improved edge detection using deep learning \cite{li2020improving}, and Borse et al. proposed InverseForm to enhance edge detection accuracy in complex images \cite{borse2021inverseform}. The recent adoption of vision transformers for recognition tasks has led to the development of dense prediction transformers specialized for semantic segmentation. Notable examples include CSWin Transformer, which captures long-range dependencies using cross-shaped windows \cite{dong2022cswin}, and HRFormer, which integrates high-resolution representations for detailed segmentation \cite{yuan2021hrformer}. These advancements have facilitated the segmentation of discrete objects and amorphous regions \cite{cheng2021per, cheng2022masked}, with transformers now incorporating token mixing via attention mechanisms \cite{guo2023visual}, multi-layer perceptron elements \cite{hou2022vision, lian2021mlp}, and pooling and convolutional blocks \cite{guo2022segnext}.

Despite setting new benchmarks in image segmentation, challenges persist, particularly under real-life conditions where RGB images are inadequate, such as in low-light environments or when capturing fast-moving subjects. Consequently, multimodal semantic segmentation is garnering increasing attention.

\subsection{Multimodality Semantic Segmentation}
Multimodal semantic segmentation is increasingly recognized for its ability to integrate diverse modal data, effectively compensating for the inherent limitations of each modality \cite{zhang2023delivering, zhang2023cmx, liang2022multimodal}. Zhou et al. \cite{zhou2022edge} integrated edge-aware features for enhanced semantic segmentation of RGB and thermal images, improving object boundary delineation in multimodal scenarios. 
Deng et al. \cite{deng2021feanet} incorporated an attention mechanism that boosts feature representation for real-time RGB-thermal semantic segmentation. Zhao et al. \cite{zhao2020didfuse} utilized deep image decomposition to fuse infrared and visible images, preserving critical features from both modalities. Huang et al. \cite{huang2022reconet} employed a recurrent network to iteratively refine multi-modality image fusion, enhancing detail preservation and reducing artifacts. 
Additionally, in RGB-depth modality fusion, ACNet \cite{hu2019acnet} employs an attention mechanism to optimize the usage of RGB and depth data for improved semantic segmentation accuracy, especially in scenarios with complex visibility. 
FuseNet \cite{hazirbas2017fusenet} integrates RGB and depth data using a dual-stream CNN, leveraging depth as an auxiliary input to enhance segmentation accuracy.
The evolution of model frameworks has progressed from those based on CNN \cite{chen2017deeplab, chen2018encoder} to those founded on Transformers \cite{xie2021segformer,guo2022segnext,dong2022cswin}. 
This transition facilitates a more nuanced analysis of the interplay between global semantics and local features, enhancing feature extraction.
For modal fusion, some techniques employ attention mechanisms to integrate different modalities \cite{hu2019acnet}. For example, CACFNet \cite{zhou2023cacfnet} utilizes cross-attention mechanisms to selectively enhance the integration of contextual information from different modalities to improve semantic correlation and feature extraction efficiency. 
Other approaches employ convolution as a feature fusion extraction module \cite{li2023residual}; Reza et al. \cite{reza2023multimodal} introduced specialized convolution layers designed to process and merge information from multiple modalities effectively. 
EGFNet \cite{dong2023egfnet} uses gated convolution to selectively fuse the most relevant features from diverse modalities, improving fusion effectiveness. 
Apart from that, Zhang et al. \cite{zhang2023delivering} innovated by adapting pooling strategies to multimodal contexts, optimizing feature reduction and abstraction processes to better accommodate the diverse characteristics of different data types.
Despite these advancements, many existing modal fusion models rely excessively on RGB images, often involving specialized feature extractors for the RGB channels. This strategy risks neglecting the variable significance of different modalities across various scenarios. To address this, we propose the Unbiased Multiscale Modal Fusion Model for Multimodal Semantic Segmentation.

\section{Method}
\begin{figure*}
\centering
\includegraphics[width=\linewidth]{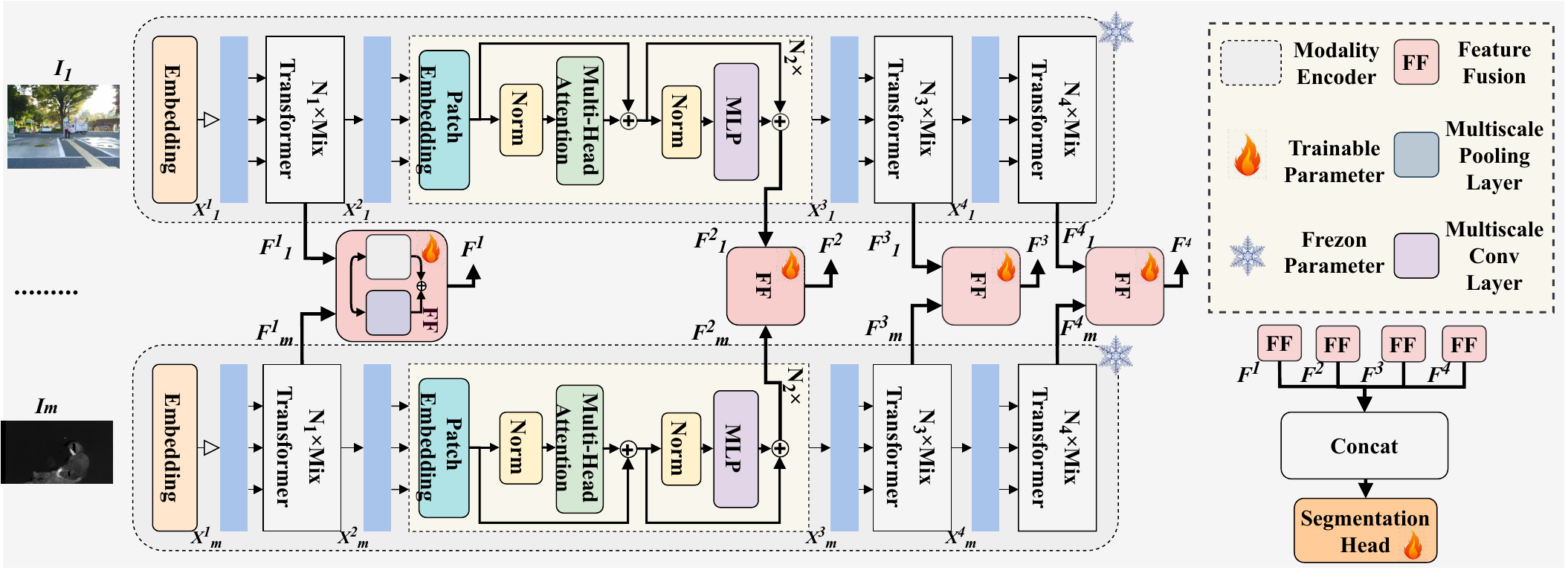}
\caption{Unbiased Multiscale Modal Fusion Model. Utilizing Segformer \cite{xie2021segformer} with frozen parameters as the feature extractor. Each modality's information is fed into respective feature extractors, divided into four distinct scales for unbiased fusion of multiscale information. Each feature fusion layer comprises two modules based on multiscale pooling and convolution, adaptively extracting features with varied scales. In the end, the multiscale information is concatenated and fed into a shared semantic segmentation head to generate segmentation results.}
\label{fig:overall}
\vspace{-10pt}
\end{figure*}

In this section, we first present the problem definition, followed by a detailed presentation of the proposed U3M. 
\subsection{Problem Definition}
Multimodal semantic segmentation seeks to assign a semantic label to each pixel in an image from predefined categories, such as 'car', 'tree', or 'road'. Distinct from traditional semantic segmentation which depends on a single modality, typically RGB images, multimodal semantic segmentation leverages data from various sources or sensors to improve the accuracy and robustness of segmentation.

Given a set \( \mathcal{I} = \{I_1, I_2, \dots, I_M\} \) of images, where \( I_m \in \mathbb{R}^{H \times W \times C_m} \) represents the \( m \)-th modality out of \( M \) total modalities. Here, \( H \), \( W \), and \( C_m \) denote the height, width, and channel size of the image, respectively. The objective of multimodal semantic segmentation is to compute a segmentation map \( S \in \mathcal{P}^{H \times W} \). Here, \( \mathcal{P} = \{1, 2, \dots, N\} \) denotes the set of \( N \) semantic labels. 

Each encoder (\(\text{Enc}_m: \mathbb{R}^{H \times W \times C_m} \rightarrow \mathbb{R}^{H' \times W' \times C_f} \)) processes the corresponding modality \( I_m \), yielding feature maps \( F^m \). These are fused by \( \text{Fusion}: (\mathbb{R}^{H' \times W' \times C_f})^M \rightarrow \mathbb{R}^{H' \times W' \times F'} \) to produce a combined representation \( F \) ($C_f$=$F'$ in this paper). Finally, \( S \) is obtained through a decoder \( \text{Dec}: \mathbb{R}^{H' \times W' \times F'} \rightarrow \mathcal{P}^{H \times W} \).

The effectiveness of a multimodal semantic segmentation model is typically evaluated based on its ability to accurately segment objects and scenes under varying conditions, making use of the additional information provided by the various modalities to overcome the limitations of single-modality segmentation.

\subsection{Overall model Architecture}
The schema of our model's architecture is presented in Fig. \ref{fig:overall} accommodating \( M \) discrete modalities. The architecture's sophistication lies in its utilization of modality-specific encoders, which are tailored to distill unique feature hierarchies from each modality, encapsulated by the equation:
\begin{equation}
F_m = \text{Enc}_m (I_m),
\end{equation}
where \( I_m \in \mathbb{R}^{H \times W \times C_m} \) represents the modality-specific input imagery for \( m \in \{1,2,\ldots,M\} \), and \( \text{Enc}_m(\cdot) \) is the corresponding encoder. These encoders are adept at generating a spectrum of feature maps at diminished resolutions—precisely, \(\frac{1}{4}, \frac{1}{8}, \frac{1}{16}, \frac{1}{32}\) of the initial resolution—collectively represented as \( F_m = \{F_{m}^1, F_{m}^2, F_{m}^3, F_{m}^4\} \). For the sake of brevity, the feature map's dimensions at the \( i \)-th encoding stage are denoted as \( (H'_i \times W'_i \times C_{fi}) \), with \( i \) traversing \{1, 2, 3, 4\}.
A quartet of fusion blocks, each allied to a corresponding encoder stage, is orchestrated to merge the features from each encoding cascade. The prominent features \( F^i_m \) from every modality are assimilated within the \( i \)-th fusion module:
\begin{equation}
F^i = \text{FusionBlock}^i(\{F^i_m\}_m).
\end{equation}
This integrative process yields a synthesized feature composite \( F = \{F^1, F^2, F^3, F^4\} \), with \( F^i \) signifying the fused feature at the \( i \)th stage. Culminating the process, these aggregated features \( F \) are input into an MLP decoder, as elucidated in \cite{xie2021segformer}, to extrapolate the segmentation contours.

\subsection{Modality Feature Encoder}
Utilizing a mix transformer encoder as referenced in \cite{xie2021segformer}, our system effectively extracts hierarchical features from various input data types. Every image \( I_m \) undergoes a patch embedding process, getting segmented into \( 4 \times 4 \) patches as per the method described in \cite{xie2021segformer}, before being processed by the mixed transformer encoder units. Illustrated in Fig. \ref{fig:overall}  is the mix transformer unit's structure, with \( X^i_{m} \) denoted the \( m \)-th modal input for the \( i \)-th mix transformer, which has the shape of \( \mathbb{R}^{H_i \times W_i \times C_i} \). This input is then reshape into a \( N_i \times C_i \) matrix, where \( N_i \) equates to \( H_i \times W_i \), for utilization as the query \( Q \), the key \( K \), and the value \( V \).
In an effort to curtail computational demands, we employ spatial reduction as suggested by \cite{xie2021segformer}, leveraging a reduction ratio \( R \). The matrices \( K \) and \( V \) undergo an initial transformation into \( \frac{N_i}{R} \times C_i \) matrices, followed by a remapping into \( \frac{N_i}{R} \times C_i \) matrices via a linear transformation process. Subsequently, a conventional Multi-Head Self-attention Mechanism (MHSA) is employed to map \( Q, K, V \) into intermediate representations as delineated by:
\begin{equation}
MHSA(Q, K, V) = \text{Concatenate}(head_1, \ldots, head_h)W^O,
\end{equation}
\begin{equation}
head_j = \text{Attention}(QW_j^Q, KW_j^K, VW_j^V).
\end{equation}
In this context, \( h \) signifies the total number of attention heads, with \( W_j^Q \), \( W_j^K \), \( W_j^V \), and \( W^O \) serving as the respective projection matrices within the spaces \( \mathbb{R}^{C_i \times d_k} \) and \( \mathbb{R}^{hd_v \times C_i} \), and \( d_k \), \( d_v \) denoting the dimensions of \( K \), and \( V \). The Attention function is characterized as:
\begin{equation}
\text{Attention}(Q, K, V) = \text{Softmax}\left( \frac{QK^T}{\sqrt{d_k}} \right)V.
\end{equation}
Here, \( Q, K \), and \( V \) correspond to the input query, key, and value matrices, respectively. This MHSA phase is succeeded by a mixing layer composed of two MLPs and a \( 3 \times 3 \) convolutional layer, which provides the necessary positional encoding within the transformer encoder to maximize segmentation efficacy as noted in \cite{xie2021segformer}. The computation within this layer is expressed as:
\begin{equation}
{X}_{\text{in}} = MHSA(Q, K, V),
\end{equation}
\begin{equation}
X_{\text{out}} = \text{MLP}(\text{GELU}(\text{Conv}_{3 \times 3}(\text{MLP}({X}_{\text{in}})))) + {X}_{\text{in}}.
\end{equation}
To conclude, an overlapping patch merge technique is applied to \( X_{\text{out}} \) follow \cite{xie2021segformer}, culminating in the synthesis of the ultimate output.
\subsection{Pyramidal Multiscale Modal Fusion Layer}
\begin{figure}
\centering
\includegraphics[width=\linewidth]{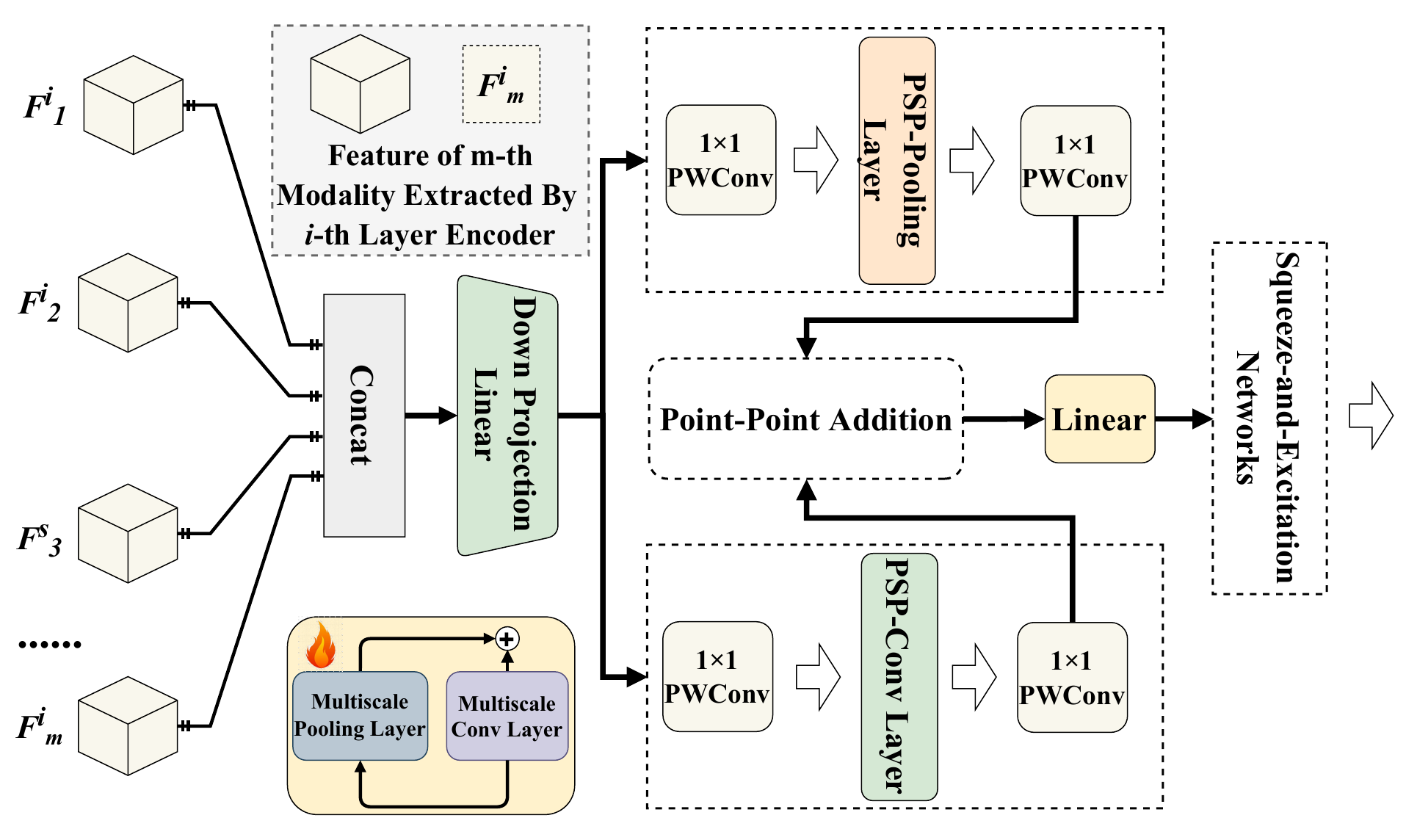}
\caption{Multiscale Feature Fusion Module. To enhance the extraction of information across multiple scales, a multiscale feature extractor is proposed.}
\label{fig:SM}
\vspace{-10pt}
\end{figure}

Feature fusion post-hierarchical extraction is performed via a designated fusion block depicted in Fig. \ref{fig:SM}. This block integrates features from modality-specific encoders across all four stages. Considering \( F^i_m \) as the input features for the \( i \)th block, where \( F^i_m \in \mathbb{R}^{H_i \times W_i \times C_i} \), sourced from \(m\)th modality, we first merge these along the channel axis, obtaining \( F'^i \). Subsequent reduction and combination of channels is achieved through a linear layer, outputting \( \tilde{F}^i \) with a reduced channel count of \( C_i \). This process is mathematically formulated as:
\begin{equation}
\tilde{F}^i = \text{Linear}(\big\Vert_{m=1}^{M} F^{i}_m).
\end{equation}
In this equation, \( \big\Vert \) signifies the concatenation of the modal features within the channel space, scaling down an \( M C_i \)-dimensional input to a \( C_i \)-dimensional output.
To enhance the extraction of information across multiple scales, a multiscale feature extractor is proposed. Specifically, we have designed two modules: the \textit{Pyramidal Convolution-Based Multimodal Information Fusion Layer} and the \textit{Pyramidal Pooling-Based Multimodal Information Fusion Layer}. They respectively utilize convolution and pooling operations across various dimensions to discern information of different granularities within the multimodal fusion features. 

\textbf{1) Pyramidal Pooling-Based Multimodal Information Fusion Layer:} The \(i\)th linear fusion feature \( \tilde{F}^i \) is refined using a multiscale feature fusion module. This unit is structured with a pair of convolutional projection layers, enclosing average pooling operations, to facilitate dimensional interlacing. Employing convolutions of sizes \(1 \times 1\), \(2 \times 2\), \(3 \times 3\), and \(6 \times 6\), the module adeptly captures and consolidates features across scales, enhancing them with residual data. Sandwiching the average pooling, the dual projection layers are instrumental in the comprehensive integration of the feature landscape.
\begin{equation}
F_{pooling}^i = \text{Conv}_{1 \times 1}(\tilde{F}^i),
\end{equation}
\begin{equation}
F_{pooling}^{i,k} = \text{Conv}_{1 \times 1}(\text{AvgPooling}_{k \times k}(F_{pooling}^i)),
\end{equation}
\begin{equation}
F_{pooling}^i = \sum_{k\in\{1,2,5,6\}}\text{Upsample}(F_{pooling}^{i,k}),
\end{equation}
\begin{equation}
F_{pooling}^i = \text{Conv}_{1 \times 1}(F_{pooling}^i ).
\end{equation}

\textbf{2) Pyramidal Convolution-Based Multimodal Information Fusion Layer:} Similar to the pooling fusion layer, the enhancement of the \(i\)th linear fusion feature \( \tilde{F}^i \) is conducted through a dedicated multiscale feature fusion apparatus. This configuration entails a sequence of convolutional projection layers that enwrap average pooling operations, thereby facilitating sophisticated feature interweaving. With convolutional kernels of dimensions \(3 \times 3\), \(5 \times 5\), and \(7 \times 7\), analogous to those employed by \cite{reza2023multimodal},
\begin{figure}
\centering
\includegraphics[width=\linewidth]{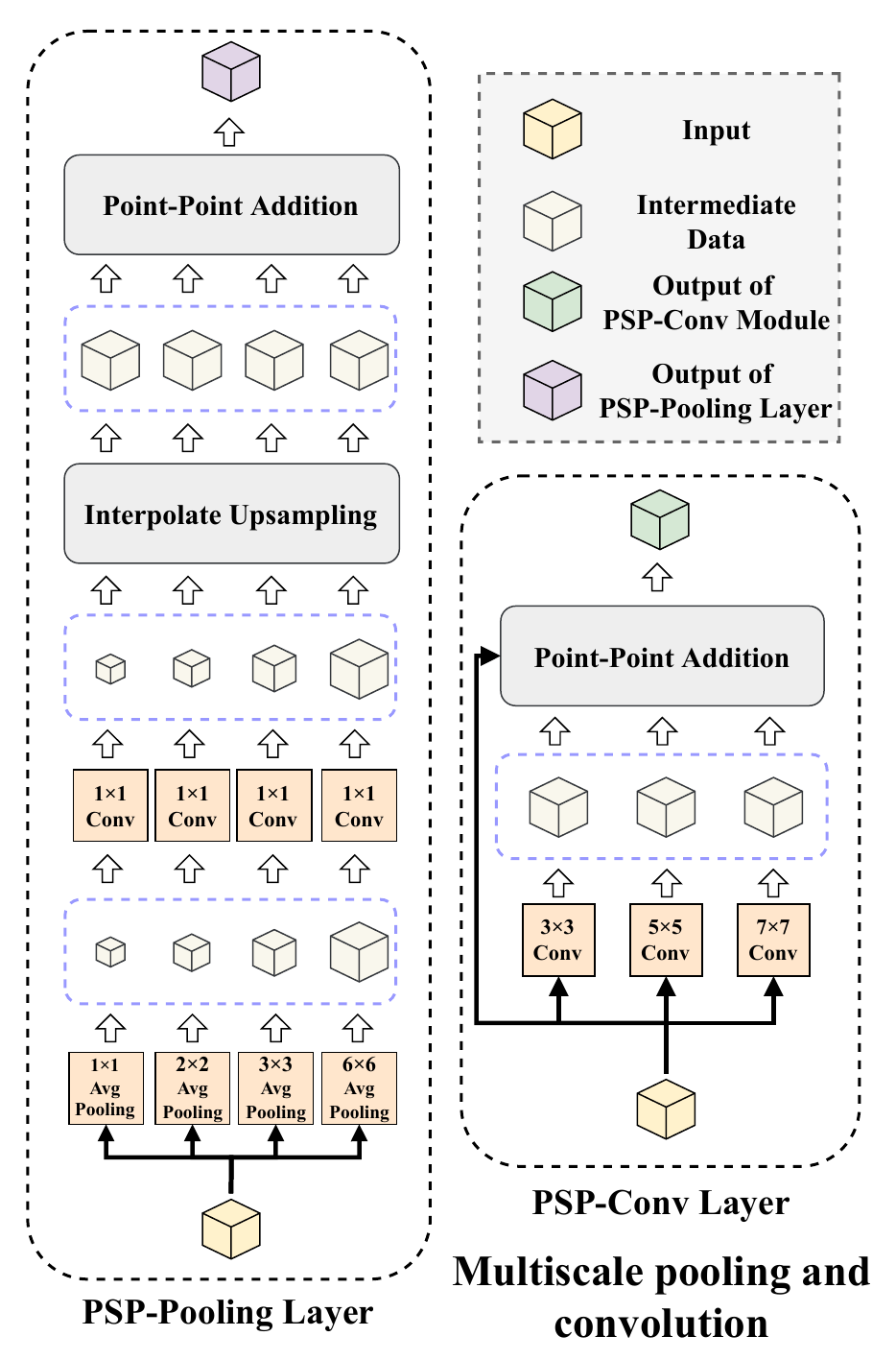}
\caption{Multiscale pooling and convolution. Pooling and convolution at different scales are capable of capturing local and global features across multiple levels, thereby complementing the global attention mechanisms integrated within the backbone architecture effectively. This approach ensures that the resultant fused features encompass a comprehensive focus on both local and global dimensions.}
\label{fig:PSPNET}
\vspace{-10pt}
\end{figure}
\begin{equation}
F_{conv}^i = \text{Conv}_{1 \times 1}(\tilde{F}^i),
\end{equation}
\begin{equation}
F_{conv}^{i,k}=\text{Conv}_{k \times k}(F_{conv}^i),
\end{equation}
\begin{equation}
F_{conv}^i = \sum_{k \in \{3,5,7\}} (F^i + F_{conv}^{i,k}),
\end{equation}
\begin{equation}
F_{conv}^i = \text{Conv}_{1 \times 1}(F_{conv}^i).
\end{equation}
The two varieties of fine-grained features extracted are fused via addition, 
\begin{equation}
F_{fusion}^i = F_{Pooling}^i + F_{conv}^i.
\end{equation}
Subsequently, $F_{fusion}^i$ passes through a linear layer and a Channel Attention Mechanism (CA) \cite{hu2018squeeze} for feature refinement. 

\subsection{Shared Segmentation Head}
The fused features generated from all the 4 fusion blocks are sent to the shared MLP decoder. We use the decoder design proposed in \cite{xie2021segformer}. The segmentation head shown in Fig. \ref{fig:overall} can be represented as the following equations:
\begin{equation}
\hat{F}^i = \text{Linear}(F_{out}^i), \quad \forall i \in \{1,2,3,4\}
\end{equation}
\begin{equation}
\hat{F}^i = \text{Upsample}(\hat{F}^i), \quad \forall i \in \{1,2,3,4\}
\end{equation}
\begin{equation}
F = \text{Linear}(\hat{F}^1 \parallel \ldots \parallel \hat{F}^i),
\end{equation}
\begin{equation}
P = \text{Linear}(F).
\end{equation}
The first linear layers take the fused features of different shapes and generate features having the same channel dimension. Then the features are up-sampled to \(\frac{1}{4}\) of the original input shape, concatenated along the channel dimension and passed through another linear layer to generate the final fused feature \( F' \). Finally, \( F \) is passed through the last linear layer to generate the predicted segmentation map \( P \).

% ===================================================================================================================================
\section{Experiments}

\subsection{Experimental Setup and Parameters}
\begin{figure*}
\centering
\includegraphics[width=\linewidth]{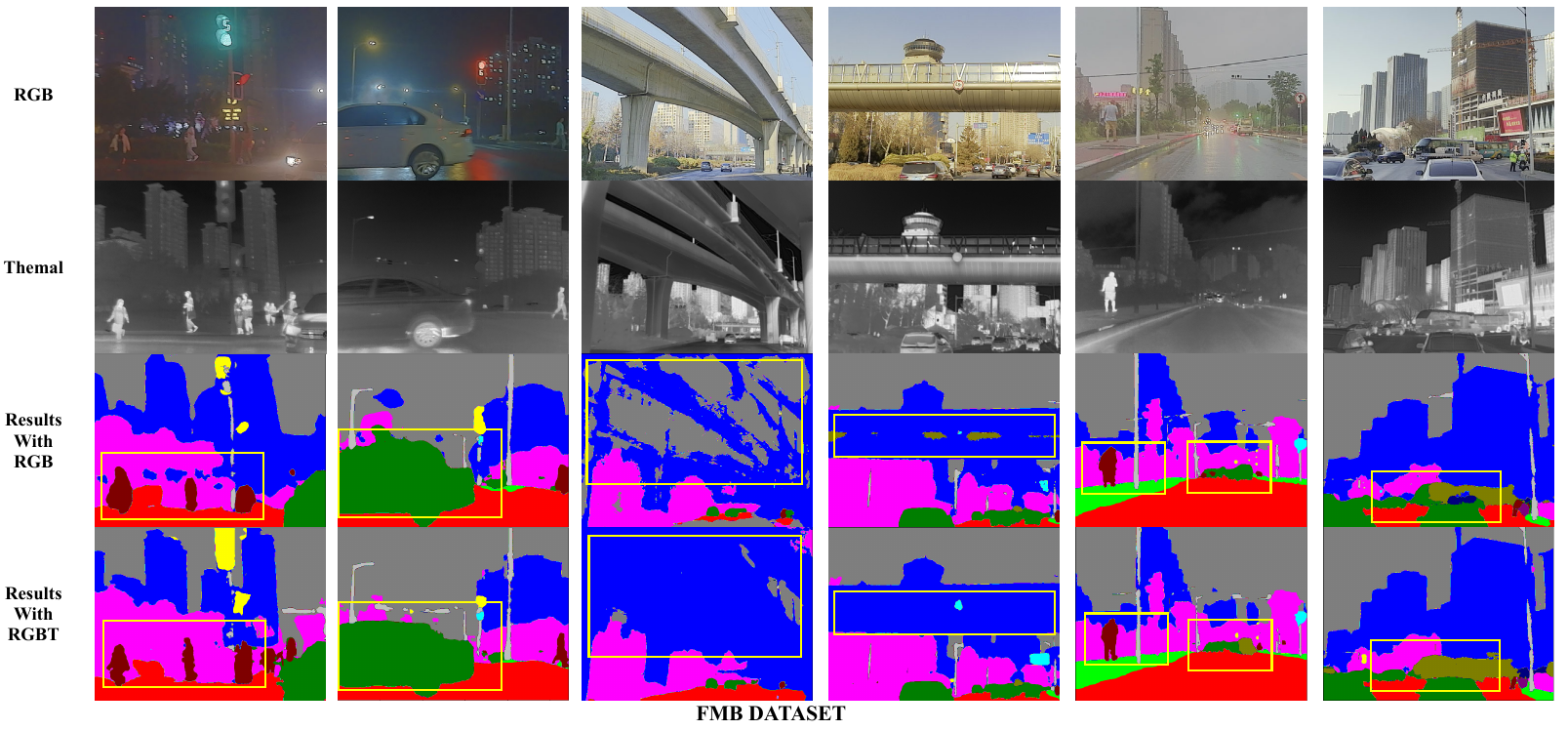}
\caption{Visualization on FMB dataset. This figure presents a comprehensive visualization of semantic segmentation results on the FMB dataset, juxtaposing RGB images with their thermal counterparts and the corresponding segmentation outcomes from two different methods: Results with RGB and Results with RGB-T. This figure exemplifies the pivotal role of integrating multimodal data in enhancing segmentation accuracy under varied environmental and lighting conditions.}
\label{fig:seg_fmb}
\vspace{-10pt}
\end{figure*}
\textbf{Experiment Configuration.} All experiments were conducted on a computational platform equipped with 4 NVIDIA GeForce RTX 4090 GPUs. To ensure the reproducibility of results, the experiments were performed under consistent hardware and software configurations.

\textbf{Parameters.} Our model utilized uniform training parameters across all datasets. The learning rate was initially set at \(6 \times 10^{-5}\), with the Adam optimizer used for adjustments. The batch size was established at 4, with a total training epoch of 400 epochs for the Mcubes dataset and 120 for the FMB dataset. Cross-entropy loss function was employed. Data augmentation techniques, including random rotations, scaling, and horizontal flipping, were applied to enhance the model's generalization capabilities.

\subsection{Dataset}

\textbf{Mcubes.} \cite{liang2022multimodal} The MCubeS dataset contains sets of RGB, Near-Infrared (NIR), Degree of Linear Polarization (DoLP), and Angle of Linear Polarization (AoLP) pairs. It is designed for researching semantic material segmentation across 20 categories. The dataset comprises 302/96/102 image pairs allocated for training/validation/testing, all sized at 1224×1024.

\textbf{FMB.} \cite{liu2023multi} The FMB dataset is a comprehensive multi-modality benchmark designed for image fusion and segmentation. It contains 1,500 well-registered pairs of infrared and visible images, annotated with 15 pixel-level categories. The training and test set contains 1220 and 280 image pairs respectively.  These images encompass a variety of real-world conditions such as dense fog and low-light scenarios, making them particularly suitable for autonomous driving and semantic understanding applications. The dataset aims to improve the generalization capabilities of fusion and segmentation models across diverse environmental conditions.
% ===================================================================================================================================
\subsection{Experimental Results}
% ===================================================================================================================================
% table-1 
\begin{table}[ht]
\centering
\caption{Performance comparison on Multimodal Material Segmentation (MCubeS) dataset \cite{liang2022multimodal}. Here A, D, and N represent angle of linear polarization (AoLP), degree of linear polarization (DoLP), and near-infrared (NIR) respectively.}
\scalebox{1.0}{
\begin{tabular}{lcc}
\hline
Method            & Modalities & \% mIoU \\
\hline
DRConv \cite{chen2021dynamic}   & RGB-A-D-N  & 34.63   \\
DDF \cite{zhou2021decoupled}     & RGB-A-D-N  & 36.16   \\
TransFuser \cite{prakash2021multi}     & RGB-A-D-N  & 37.66   \\
DeepLabv3+ \cite{chen2018encoder}     & RGB-A-D-N  & 38.13      \\
MMTM \cite{joze2020mmtm}     & RGB-A-D-N  &  39.71   \\
FuseNet \cite{hazirbas2017fusenet}     & RGB-A-D-N  & 40.58   \\
MCubeSNet \cite{liang2022multimodal}     & RGB-A-D-N  &  \textcolor{green}{42.46}   \\
CMNeXt \cite{zhang2023delivering}     & RGB-A-D-N  &  \textcolor{blue}{51.54}   \\
\hline
U3M (Ours)  & RGB-A-D-N  & \textcolor{red}{51.69}   \\
\hline
\end{tabular}}
\label{tab:mcubes_performance}
\end{table}

\textbf{1) Results on Mcubes Dataset.}
Table \ref{tab:mcubes_performance} compares different methods based on mean Intersection over Union (mIoU) percentage, which is a common metric for evaluating the accuracy of segmentation models. The methods include various modalities, specifically RGB-A-D-N, where A, D, and N represent angle of linear polarization (AoLP), degree of linear polarization (DoLP), and near-infrared (NIR) respectively. Our model achieves the highest mIoU of 51.69, significantly outperforming all other listed methods. This suggests that U3M is particularly effective at handling multimodal inputs to provide superior segmentation accuracy. The closest competitor, CMNeXt, achieves a mIoU of 51.54, making our model's lead relatively narrow but still notable. The superior performance of U3M suggests its potential applicability in real-world scenarios where precise material identification is crucial, such as in autonomous driving environments or quality control in manufacturing. 

The mIoU results of each category from the MCubeS dataset provide a nuanced view of the segmentation performance across diverse material categories, as shown in Table \ref{tab:per_class_iou}. Our proposed model demonstrates robust performance, often surpassing state-of-the-art models in per-class IoU metrics. Notably, our model achieves superior results in classes such as 'Leaf' (76.4), 'Water' (63.8), and 'Cobblestone' (73.5), indicating its proficiency in handling complex textures and varied lighting conditions that these materials present. However, some categories like 'Human' (12.2) and 'Plastic' (26.3) exhibit weaker performance, suggesting potential areas for further model refinement. This underperformance could be attributed to the challenges associated with the high variability in human appearances and the often subtle differences in plastic materials' visual characteristics under different conditions. Comparatively, our model's performance in the 'Rubber' (71.7) and 'Leaf' (76.4) categories is particularly noteworthy, underscoring its effectiveness in segmenting materials with distinct textural properties. The overall mean IoU of 51.7 places our model competitively within the landscape of current methods, closely following the CMNeXt model, which exhibits a slightly lower mean IoU of 51.5.
% % ----------------------- table-2
\begin{table*}[ht]
\centering
\caption{Per-class \% IoU comparison on MCubeS dataset. Our proposed model shows better performance in detecting most of the classes compared to the current state-of-the-art models. * indicates that the code and pretrained model from the authors were used to generate the results.}
\label{tab:per_class_iou}
\resizebox{\textwidth}{!}{%
\begin{tabular}{lcccccccccccccccccccccc}
\hline
Methods     & Asphalt & Concrete & Metal & Road marking & Fabric & Glass & Plaster & Plastic & Rubber & Sand & Gravel & Ceramic & Cobblestone & Brick & Grass & Wood & Leaf & Water & Human & Sky & Mean \\
\hline
MCubeSNet \cite{liang2022multimodal} & \textcolor{blue}{85.7} & 42.6 & 47.0 & 59.2 & 12.5 & 44.3 & \textcolor{red}{3.0} & 10.6 & 12.7 & 66.8 & \textcolor{blue}{67.1} & \textcolor{red}{27.8} & 65.8 & 36.8 & 54.8 & 39.4 & 73.0 & 13.3 & 0.0 & 94.8 & 42.9 \\
CMNeXt \cite{zhang2023delivering}*   & 84.3 & \textcolor{red}{44.9} & \textcolor{blue}{53.9} & \textcolor{red}{74.5} & 32.3 & \textcolor{red}{54.0} & 0.8 & \textcolor{red}{28.3} &  \textcolor{red}{29.7} & \textcolor{blue}{67.7} & 66.5 & \textcolor{blue}{27.7} & \textcolor{blue}{68.5} & \textcolor{blue}{42.9} & \textcolor{blue}{58.7} & \textcolor{red}{49.7} & \textcolor{blue}{75.4} & \textcolor{blue}{55.7} & \textcolor{red}{18.9} & \textcolor{red}{96.5} & \textcolor{blue}{51.5} \\
\hline
(Ours)                               & \textcolor{red}{86.2} & \textcolor{blue}{44.5} & \textcolor{red}{55.1} & \textcolor{blue}{68.0} & \textcolor{red}{33.4} & \textcolor{blue}{53.9} & \textcolor{blue}{1.2} & \textcolor{blue}{26.3} & 26.9 & \textcolor{red}{68.2} & \textcolor{red}{71.7} & 25.0 & \textcolor{red}{73.5} & \textcolor{red}{44.1} & \textcolor{red}{59.7} & \textcolor{blue}{47.4}  & \textcolor{red}{76.4} & \textcolor{red}{63.8} & \textcolor{blue}{12.2} & \textcolor{blue}{96.4}  & \textcolor{red}{51.7} \\
\hline
\end{tabular}%
}
\end{table*}

% % -----------------------table-3
\begin{table}[ht]
\centering
\caption{Performance comparison on FBM dataset \cite{liu2023multi}. We show performance for different methods from already published works.}
\begin{tabular}{lcc}
\hline
Methods            & Modalities  & \% mIoU \\
\hline
GMNet \cite{zhou2021gmnet}    & RGB-Infrared & 49.2    \\
LASNet \cite{li2022rgb}   & RGB-Infrared & 42.5    \\
EGFNet \cite{zhou2022edge}   & RGB-Infrared & 47.3    \\
FEANet \cite{deng2021feanet}   & RGB-Infrared & 46.8    \\
DIDFuse \cite{zhao2020didfuse}   & RGB-Infrared & 50.6    \\
ReCoNet \cite{huang2022reconet}   & RGB-Infrared & \textcolor{green}{50.9}    \\
U2Fusion \cite{xu2020u2fusion}   & RGB-Infrared & 47.9    \\
TarDAL \cite{liu2022target}   & RGB-Infrared &  48.1    \\
SegMiF \cite{liu2023multi}   & RGB-Infrared & \textcolor{blue}{54.8}    \\
\hline
U3M (Ours)   & RGB-Infrared & \textcolor{red}{60.8}    \\
\hline
\end{tabular}
\label{tab:fbm_performance}
\end{table}

% ----------------------- table-4
\begin{table*}[ht]
\centering
\caption{Per-class \% IoU comparison on FMB \cite{liu2023multi} dataset for both RGB only and RGB-infrared modalities. T-Lamp and T-Sign stand for Traffic Lamp and Traffic Sign respectively. Our model outperforms all the methods for all the classes except for the truck class.}
\label{tab:per_class_iou_fmb}
\scalebox{1.0}{
\begin{tabular}{lcccccccccc}
\hline
Methods & Modalities & Car & Person & Truck & T-Lamp & T-Sign & Building & Vegetation & Pole & \% mIoU \\
\hline
GMNet \cite{zhou2021gmnet} & RGB-Infrared       & 79.3 & 60.1 & 22.2 & 21.6 & 69.0 & 79.1 & 83.8 & 39.8 & 49.2 \\
LASNet \cite{li2022rgb} & RGB-Infrared          & 72.6 & 48.6 & 14.8 & 2.9 & 59.0 & 75.4 & 81.6 & 36.7 & 42.5 \\
EGFNet \cite{zhou2022edge} & RGB-Infrared       & 77.4 & 63.0 & 17.1 & 25.2 & 66.6 & 77.2 & 83.5 & 41.5 & 47.3 \\
FEANet \cite{deng2021feanet} & RGB-Infrared     & 73.9 & 60.7 & 32.3 & 13.5 & 55.6 & 79.4 & 81.2 & 36.8 & 46.8 \\
DIDFuse \cite{zhao2020didfuse} & RGB-Infrared   & 77.7 & 64.4 & 28.8 & 29.2 & 64.4 & 78.4 & 82.4 & 41.8 & 50.6 \\
ReCoNet \cite{huang2022reconet} & RGB-Infrared  & 75.9 & \textcolor{blue}{65.8} & 14.9 & 34.7 & 66.6 & 79.2 & 81.3 & 44.9 & 50.9 \\
U2Fusion \cite{xu2020u2fusion} & RGB-Infrared   & 76.6 & 61.9 & 14.4 & 6.8 & 68.9 & 78.8 & 82.2 & 42.2 & 47.9 \\
TarDAL \cite{liu2022target} & RGB-Infrared      & 74.2 & 56.0 & 18.3 & 7.8 & 69.0 & 79.1 & 81.7 & 41.9 & 48.1 \\
SegMiF \cite{liu2023multi} & RGB-Infrared       & 78.3 & 65.4 & 18.8 & 6.5 & 64.8 & 78.0 & 85.0 & \textcolor{red}{49.8} & 54.8 \\
\hline
(Ours) & RGB                           & \textcolor{red}{83.3} & 57.6 & \textcolor{blue}{41.6} & \textcolor{blue}{42.7} & \textcolor{blue}{78.3} & \textcolor{red}{81.7} & \textcolor{blue}{85.6} & \textcolor{blue}{49.5} & \textcolor{blue}{60.5} \\
(Ours) & RGB-Infrared                           & \textcolor{blue}{82.3} & \textcolor{red}{66.0} & \textcolor{red}{41.9} & \textcolor{red}{46.2} & \textcolor{red}{81.0} & \textcolor{blue}{81.3} & \textcolor{red}{86.8} & 48.8 & \textcolor{red}{60.8} \\
\hline
\end{tabular}}
\end{table*}

% ----------------------- TABLE-4
\begin{table}[ht]
\centering
\caption{Performance evaluation (measured in \% mIoU) on the Multimodal Material Segmentation (MCubeS) dataset \cite{liang2022multimodal} across various modality pairings is presented. The modalities A, D, and N correspond to angle of linear polarization (AoLP), degree of linear polarization (DoLP), and near-infrared (NIR), respectively.}
\label{tab:performance_comparison_different_modalities}
\scalebox{1.0}{
\begin{tabular}{lccc}
\toprule
Modalities & MCubeSNet \cite{liang2022multimodal} & CMNeXt \cite{zhang2023delivering} & (Ours) \\
\midrule
RGB        & 33.70               & 48.16            & \textcolor{red}{49.22}            \\
RGB-A      & 39.10               & 48.42            & \textcolor{red}{49.89}             \\
RGB-A-D    & 42.00               & 49.48            & \textcolor{red}{50.26}             \\
RGB-A-D-N  & 42.86               & 51.54            & \textcolor{red}{51.69}             \\
\bottomrule
\end{tabular}%
}
\end{table}

% % -----------------------table-6
\begin{table}[ht]
\centering
\caption{Performace comparison on Mcubes dataset with different combinations of proposed modules \cite{liu2023multi}.}
\scalebox{1.0}{
\begin{tabular}{lcc}
\hline
Methods            & Modalities  & \% mIoU \\
\hline
Linear Layer Only   & RGB-ADN & 49.89    \\
Linear Layer + ChannelAttention   & RGB-ADN & 50.34    \\
Linear Layer + PSPModule   & RGB-ADN & 50.62   \\
U3M   & RGB-ADN & \textcolor{red}{51.69}    \\
\hline
\end{tabular}}
\label{tab:modules_combination_performance}
\end{table}

\textbf{2) Results on FMB Dataset.}
The results presented in Table \ref{tab:fbm_performance} for the performance comparison on the FBM dataset highlight the capabilities of various semantic segmentation models utilizing RGB-Infrared modalities, a combination critical for enhancing material differentiation under varying illumination conditions. Notably, our model, U3M, achieves an impressive mIoU score of 60.8, which surpasses all other models listed. This superior performance can be attributed to the effective integration of RGB and infrared data, which allows U3M to robustly capture and utilize the complementary information provided by these modalities. Infrared imaging, known for its utility in low-light conditions and its ability to differentiate materials based on thermal properties, combined with the rich detail available in RGB images, provides a more comprehensive understanding of the scene, enhancing segmentation accuracy. The closest competitors, SegMif and TarDAL, achieve mIoU scores of 54.8 and 48.1, respectively, which indicates that while these models are effective, there remains a significant gap in performance compared to U3M. This gap suggests that U3M may employ more sophisticated or optimized techniques for multimodal integration, perhaps through advanced feature fusion strategies or more effective neural network architectures.

Table \ref{tab:per_class_iou_fmb} provides a detailed per-class percentage of mIoU analysis for models tested on the FMB dataset, utilizing both RGB and RGB-infrared modalities. The data demonstrates the superior performance of our model across the majority of categories, with particularly high IoU scores in 'Building' (81.3) and 'Vegetation' (86.8), illustrating its robustness in identifying and segmenting structural and natural elements in urban environments. However, a notable exception is the 'Truck' category, where our model's performance (41.9) lags behind other models like the HDNet (60.7) and EGNet (63.0), indicating potential challenges in distinguishing larger vehicles possibly due to their similar spectral signatures with other objects or insufficient training data representing this category. Our model's overall mIoU of 60.8 is the highest among the listed methods, confirming its efficacy in integrating RGB and infrared data to enhance segmentation accuracy. The integration of infrared data is particularly beneficial in improving the model’s performance under varying lighting conditions, as infrared provides consistent material recognition capabilities that are less susceptible to variations in visible light. The segmentation performance in the 'Traffic Lamp' (46.2) and 'Traffic Sign' (62.2) classes, while not the highest, still represents competitive results, suggesting that the model effectively utilizes the infrared component to detect these objects typically characterized by their distinct material properties not always apparent in RGB imagery.
\begin{figure*}
\centering
\includegraphics[width=\linewidth]{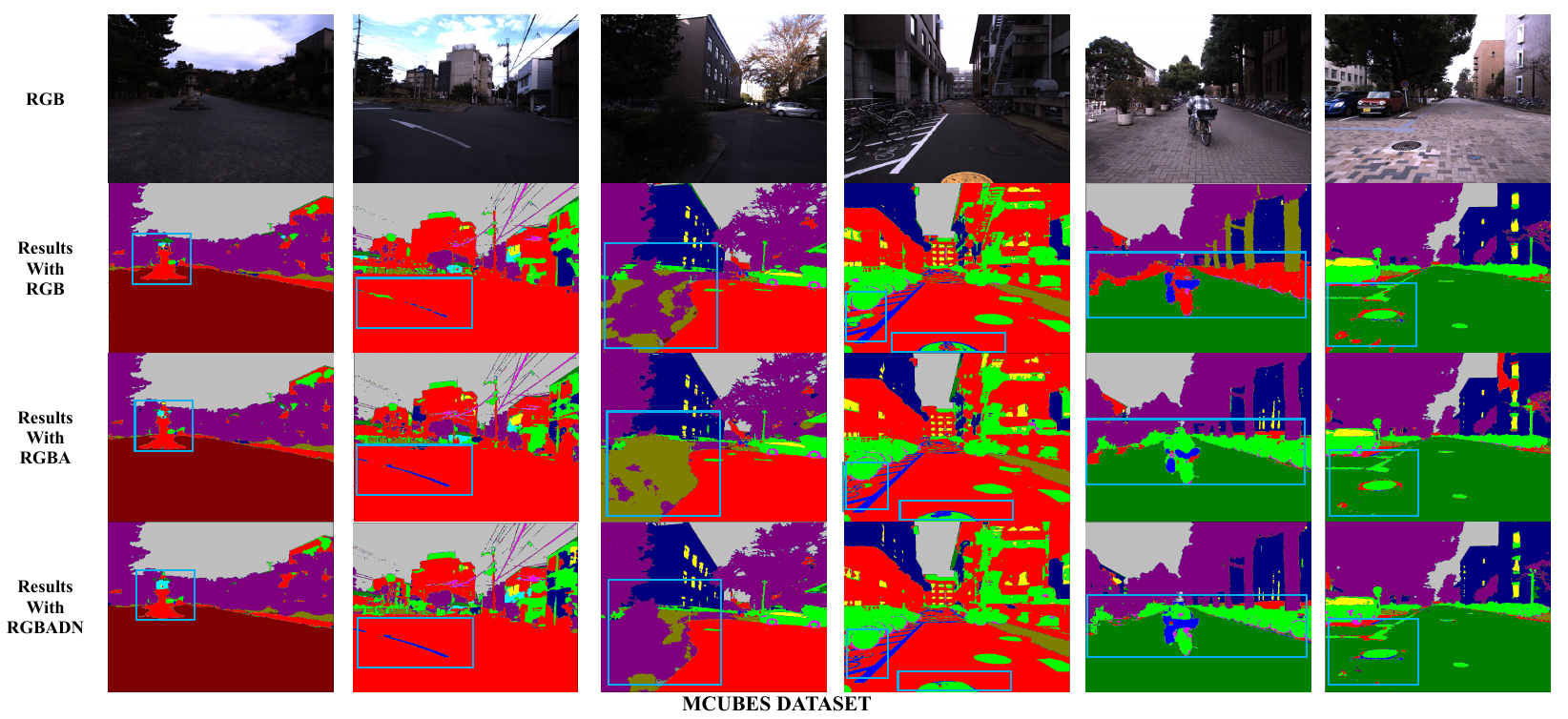}
\caption{Segmentation results visualization on Mcubes dataset. The figure provides a vivid visualization of semantic segmentation results on the Mcubes dataset, showcasing the effectiveness of different modalities: RGB, RGB-A, RGB-A-D, and RGB-A-D-N.}
\label{fig:seg_mcubes}
\vspace{-10pt}
\end{figure*}

\subsection{Visulization}

\textbf{1) Segmentation result visualization on FMB dataset}. The visualization in Figure 6 compellingly illustrates the effectiveness of multimodal semantic segmentation using RGB and thermal (RGB-T) data on the FMB dataset. The comparative analysis of RGB-only versus RGB-T segmentation underscores the limitations of relying solely on visible light data, particularly under adverse lighting conditions. The enhanced segmentation accuracy achieved with RGB-T highlights the thermal modality's critical role in distinguishing and classifying various elements within urban scenes, reinforcing the necessity for incorporating multimodal data in segmentation tasks to tackle diverse environmental challenges. This analysis confirms the hypothesis that thermal data significantly boosts the segmentation capability, particularly in detecting living entities and differentiating them from inanimate backgrounds, thereby providing a more reliable and robust segmentation framework for real-world applications.
\begin{figure*}
\centering
\includegraphics[width=\linewidth]{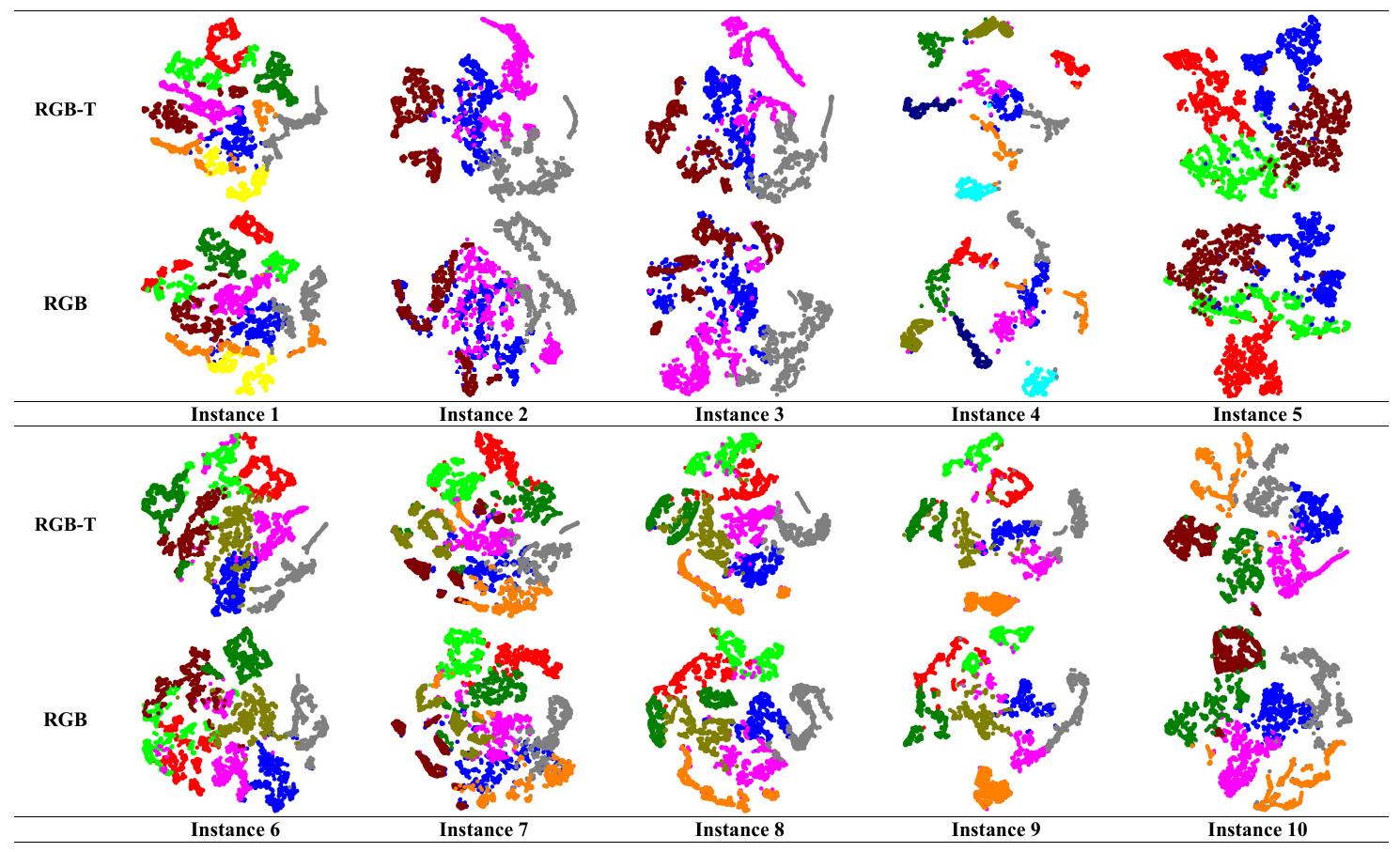}
\caption{t-SNE visualization on FMB dataset. The visualization comprises multiple panels, each line represents a different combination of modalities used for segmentation, labeled as RGB and RGBT. These panels help illustrate the distribution and separation of features in a two-dimensional space, giving insights into the discriminative power of the features extracted under different input conditions.}
\label{fig:tsne}
\vspace{-10pt}
\end{figure*}
\textbf{2) t-SNE visualization on FMB dataset}. The t-SNE visualizations in Fig. \ref{fig:tsne} underscore the significant impact of integrating thermal imagery with RGB data on the feature separability and overall effectiveness of semantic segmentation models. The enhanced separation and definition of clusters with RGB-T data corroborate the hypothesis that thermal information aids in resolving ambiguities encountered with RGB-only data, especially in complex segmentation scenarios. This analysis not only highlights the value of multimodal approaches in improving semantic segmentation tasks but also emphasizes the need for models that can effectively leverage diverse data types to achieve more accurate and robust segmentation results. Such insights are critical for advancing the development of segmentation technologies for applications in areas such as autonomous driving, environmental monitoring, and urban planning, where accurate material and object differentiation are paramount.

\textbf{3) Segmentation result visualization on Mcubes dataset}. The segmentation visualizations in Fig. \ref{fig:seg_mcubes} underscore the incremental benefits of integrating multiple data modalities into semantic segmentation tasks on the Mcubes dataset. Each added modality—AoLP, DoLP, and NIR—contributes uniquely to the enhancement of the segmentation accuracy, improving specific challenges encountered with the RGB-only model. The results vividly demonstrate that while RGB data provides a foundational layer of visual information, the incorporation of polarization and thermal data crucially enriches the feature set available for segmentation, thereby enabling more precise and contextually aware delineations of urban scene elements. This multimodal approach illustrates the potential for such technologies to be applied in real-world scenarios where diverse environmental factors and varied object interactions complicate the accurate interpretation of urban spaces.

\subsection{Ablation Experiment}
\textbf{1) Ablation experiment on different modalities}. The data presented in Table \ref{tab:performance_comparison_different_modalities} showcases the mIoU percentages for semantic segmentation performance on the Multimodal Material Segmentation (MCubeS) dataset. The analysis compares the performance of various models utilizing different combinations of modalities: RGB, RGB-A, RGB-A-D, and RGB-A-D-N. Our model demonstrates a progressive improvement in segmentation accuracy as additional modalities are integrated. Starting with RGB alone, our model achieves a mIoU of 49.22, which is higher compared to MCubeSNet's 33.70 and CMNeXt's 48.16. This trend of superior performance continues with the inclusion of AoLP (RGB-A), where our model scores 49.89, slightly ahead of CMNeXt's 48.42. The enhancement is more pronounced in configurations involving both AoLP and DoLP (RGB-A-D), with our model achieving a mIoU of 50.26, compared to CMNeXt's 49.48. The most comprehensive modality combination, incorporating RGB, AoLP, DoLP, and NIR (RGB-A-D-N), allows our model to reach its peak performance at a mIoU of 51.69, which stands out significantly against CMNeXt's 51.54 and MCubeSNet's 42.86. The incremental improvements observed with each additional modality underscore the efficacy of integrating multimodal data to capture a richer, more comprehensive feature set for material segmentation. The integration of NIR, in particular, appears to provide critical enhancements, likely due to its capability to offer consistent material properties detection that is less dependent on visible light conditions.

\textbf{2) Ablation experiment on different module combinations}. The data presented in Table \ref{tab:modules_combination_performance} illustrates a comparative analysis of semantic segmentation performance on the MCubes dataset using different architectural enhancements within a given framework. These results highlight the influence of various module integrations on the mIoU metric. Starting with a baseline configuration that utilizes a Linear Layer only, the model achieves a mIoU of 49.39. This setup serves as a foundational benchmark for evaluating the effectiveness of additional modules. Upon incorporating a Channel Attention mechanism, there is a modest increase in performance, with the mIoU improving to 50.34. Channel Attention likely aids the model in focusing more effectively on relevant features by re-weighting channel-specific features, thus providing a more refined feature map for segmentation tasks. A more substantial improvement is observed when a Pyramid Scene Parsing (PSP) module is added, resulting in the mIoU of 50.62. The PSP module, known for its capability to aggregate context information at different scales, evidently enhances the model's ability to capture and integrate multi-scale contextual information, which is crucial for accurate segmentation. Our model, U3M, which incorporates RGB-ADN modalities, achieves the highest performance with a mIoU of 51.69. This indicates that the combination of RGB data along with AoLP and DoLP from the ADN modality, effectively utilized in U3M, significantly contributes to the segmentation accuracy. This performance underscores the model’s robustness and its enhanced capability to discriminate between material types and conditions in a complex dataset like MCubes.

\section{Conclusion and Limitation}
\textbf{1) Conclusion}. The proposed U3M makes a significant leap in multimodal semantic segmentation, featuring innovative modality integration and feature fusion techniques. It addresses modal bias by employing an unbiased multiscale modal fusion methodology that equitably treats all modalities, thereby reducing manual bias. This model utilizes multiscale fusion modules that combine convolutional and pooling strategies, effectively integrating modalities at various scales and enhancing adaptability across diverse environments. Extensive experiments on challenging datasets show U3M consistently surpasses existing models in both accuracy and robustness, proving its suitability for real-world applications such as autonomous driving and urban planning.

\textbf{2) Limitation and future work}. Future work could focus on optimizing the model’s architecture for greater efficiency. Additionally, integrating more varied modalities and extensive real-world testing could broaden its application scope, ensuring it meets the evolving demands of practical implementations in areas like autonomous driving and urban planning.

% \section*{Acknowledgment}

% %Dr. Reveryrand would like to acknowledge the funding by XLIM, Limoges, France. 
% The authors would like to thank Dr. David Root and Dr. Jean-Pierre Teyssier at Agilent Technologies for the loan of the time-domain nonlinear measurement equipment and TriQuint Semiconductor for the donation of the transistors. 

% or


% ============================================
%\appendices
%Appendix one text goes here %\cite{Roberg2010}.

%\section{}

%\section*{Acknowledgment}

%The authors would like to thank D. Root for the loan of the SWAP. The SWAP that can ONLY be usefull in Boulder...

\ifCLASSOPTIONcaptionsoff
  \newpage
\fi



%\begin{thebibliography}{1}


\bibliographystyle{IEEEtran}
\bibliography{IEEEabrv}

\vfill


\end{document}